%% file: main.tex
\documentclass[letterpaper]{article} 
\usepackage{aaai2026}  
\usepackage{times}  
\usepackage{helvet}  
\usepackage{courier}  
\usepackage[hyphens]{url}  
\usepackage{graphicx} 
\usepackage{natbib}  
\usepackage{caption} 
\usepackage[linesnumbered,ruled,vlined]{algorithm2e}
\usepackage{amsmath}
\usepackage{amssymb}

\usepackage{newfloat}
\usepackage{arydshln}
\usepackage{listings}
\usepackage[dvipsnames]{xcolor}
\usepackage{xpatch}
\usepackage{multirow}
\usepackage{graphicx}
\usepackage{booktabs}
\usepackage{amssymb}
\usepackage{bbding}
\usepackage{pifont}
\usepackage{colortbl} 
\usepackage{amsmath}
\usepackage{booktabs}       

\DeclareCaptionStyle{ruled}{labelfont=normalfont,labelsep=colon,strut=off} 
\title{Can Protective Watermarking Safeguard the Copyright of 3D Gaussian Splatting?}
\author{
      Wenkai Huang \textsuperscript{\rm 1,2}\equalcontrib,
      Yijia Guo \textsuperscript{\rm 3}\equalcontrib,
      Gaolei Li \textsuperscript{\rm 1,2}\thanks{Corresponding authors.},
      Lei Ma \textsuperscript{\rm 3,4}\footnotemark[2],
      Hang Zhang \textsuperscript{\rm 5}\\
      Liwen Hu \textsuperscript{\rm 3},
      Jiazheng Wang \textsuperscript{\rm 6},
      Jianhua Li \textsuperscript{\rm 1,2}\footnotemark[2],
      Tiejun Huang \textsuperscript{\rm 3}\\
}
\affiliations{
    \textsuperscript{\rm 1}School of Computer Science, Shanghai Jiao Tong University\\
    \textsuperscript{\rm 2}Shanghai Key Laboratory of Integrated Administration Technologies for Information Security, Shanghai Jiao Tong University\\
    \textsuperscript{\rm 3}National Key Laboratory for Multimedia Information Processing, Peking University\\
    \textsuperscript{\rm 4}National Biomedical Imaging Center, Peking University \\

    \textsuperscript{\rm 5}Cornell University \\
    \textsuperscript{\rm 6}Hunan University \\
    \{sjtuhwk, gaolei\_li, lijh888\}@sjtu.edu.cn, \{2301112015, leima, liwenhu, tjhuang\}@stu.pku.edu.cn, hz459@cornell.edu, wjiazheng@hnu.edu.cn
}

\usepackage{bibentry}

\makeatletter
\newcommand{\thickhline}{%
    \noalign {\ifnum 0=`}\fi \hrule height 1pt
    \futurelet \reserved@a \@xhline
}
\makeatother
\begin{document}

\maketitle

\input{sec/0_abstract}


\input{sec/1_intro}
\input{sec/2_Relatedwork}
\input{sec/4_method}

\input{sec/5_Experiment}

\section{Acknowledgments}
This work is supported by National Nature Science Foundation of China under Grant No. 62572314, 62202303, 62471301, and 62088102, National Science and Technology Major Project (Grant No. 2022ZD0116305), and the Beijing Natural Science Foundation (Grant Nos. F251020 and JQ24023).
\bibliography{aaai2026}

\end{document}

%% file: sec/0_abstract.tex
\begin{figure*}

\centering
\includegraphics[width=\linewidth]{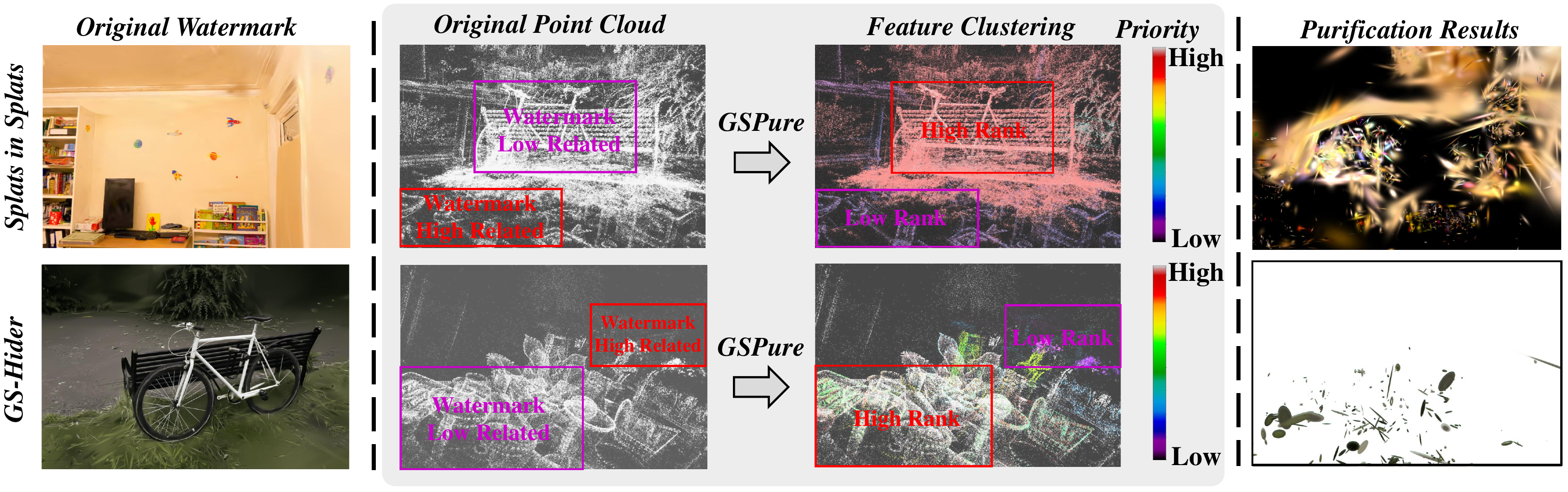}
\caption{\textbf{Visualization of purification results by \textit{GSPure} on existing scene-level watermarking techniques \cite{guo2024splats, zhang2024gshider}}. From left to right: watermark extracted by the copyright owner from the original published point cloud; the original published point cloud containing watermark information; feature clustering results which highlights watermark-correlated Gaussian primitives achieved by our GSPure (lower ranks indicate lower preservation priority); and the watermark extracted after applying our adaptive pruning. Clearly, the majority of watermark information is effectively removed.} 
\label{fig1}
\end{figure*}
\begin{abstract}
3D Gaussian Splatting (3DGS) has emerged as a powerful representation for 3D scenes, widely adopted due to its  exceptional efficiency and high-fidelity visual quality. Given the significant value of 3DGS assets, recent works have introduced specialized watermarking schemes to ensure copyright protection and ownership verification. However, can existing 3D Gaussian watermarking approaches genuinely guarantee robust protection of the 3D assets? In this paper, for the first time, we systematically explore and validate possible vulnerabilities of 3DGS watermarking frameworks. We demonstrate that conventional watermark removal techniques designed for 2D images do not effectively generalize to the 3DGS scenario due to the specialized rendering pipeline and unique attributes of each gaussian primitives. Motivated by this insight, we propose \textit{\textbf{GSPure}}, the first watermark purification framework specifically for 3DGS watermarking representations. By analyzing view-dependent rendering contributions and exploiting geometrically accurate feature clustering, GSPure precisely isolates and effectively removes watermark-related Gaussian primitives while preserving scene integrity. Extensive experiments demonstrate that our GSPure achieves the best watermark purification performance, reducing watermark PSNR by up to 16.34dB while minimizing degradation to original scene fidelity with less than 1dB PSNR loss. Moreover, it consistently outperforms existing methods in both effectiveness and generalization. Our code is available at \url{https://github.com/insightlab-CG-3DV/GSPure}.

\end{abstract}

%% file: sec/1_intro.tex
\section{Introduction}
\label{sec:intro}
3D Gaussian Splatting (3DGS) has recently become a prominent method for efficiently representing and rendering complex 3D scenes \cite{3dgs}. By encoding scenes through anisotropic Gaussian primitives, it achieves both high-fidelity visual results and computational efficiency, driving its rapid adoption in generative tasks like novel-view synthesis \cite{paliwal2024coherentgs}, 3D modeling \cite{feng2025flashgs, sun20243dgstream}, and virtual reality \cite{tu2025vrsplat}. However, as 3DGS gains popularity, its high training cost makes the models valuable digital assets \cite{zhang2024gshider}. This raises critical concerns regarding their security and ownership, particularly in preventing unauthorized distribution or tampering that could lead to intellectual property violations. To mitigate these risks, researchers are developing specialized watermarking techniques for 3D Gaussian Splatting to enable reliable asset identification, ownership verification, and provenance tracking \cite{guo2024splats, zhang2024gshider, jang20243d, zhang2025securegs}.

Traditional 2D watermarking techniques are highly vulnerable to purification and detection attacks~\cite{li2021neural, wang2024robust, wu2021adversarial}, significantly diminishing their effectiveness. However, these neural network-based or 2D image transformation-based watermark purification methods pose no direct threat to existing 3D Gaussian watermarking schemes, as they inherently overlook the geometric continuity and intrinsic properties of 3DGS. Moreover, current 3D Gaussian watermarking frameworks are carefully designed to embed watermark information into the model itself rather than merely into rendered images~\cite{guo2024splats, zhang2024gshider}. \textbf{Therefore, we primarily focus on watermarking methods that target 3D scene hiding, as traditional 2D watermark purification strategies exhibit poor generalization and limited transferability when applied to embedded 3D scenes.} 

Motivated by the limitations, we present \textbf{\textit{GSPure}}, the first dedicated watermark purification framework specifically designed for 3DGS. Unlike existing methods that operate exclusively in the rendered image domain, GSPure exploits the intrinsic geometric and rendering properties of 3D Gaussian primitives. Particularly, we observe that watermark-related Gaussian primitives exhibit distinct viewpoint-dependent rendering characteristics compared to those associated with the original scene. Based on this insight, we propose a \textbf{\textit{view-aware gaussian weight accumulation}} strategy to quantify the rendering contributions of original-scene Gaussians across multiple viewpoints. To further enhance the accuracy of Gaussian separation, we introduce \textbf{\textit{geometrically accurate feature clustering}}, which integrates Gaussian positions, opacity, and computed contributions into a unified geometric feature space. By applying adaptive clustering, GSPure effectively isolates and removes watermarks while preserving original scene fidelity. Comprehensive experiments demonstrate that GSPure achieves superior watermark purification performance across existing scene hiding methods, highlighting its robustness, generalization, and practical effectiveness.
Our contributions are three-fold:

\begin{itemize}
\item
We are the first to systematically explore watermark purification techniques for 3DGS watermarking and propose \textit{GSPure}, the first framework designed specifically for purifying 3D scene-level watermarks.

\item
To accurately isolate watermarks, we propose a view-aware gaussian weight accumulation mechanism to quantify Gaussian contributions. Additionally, we introduce a geometrically accurate feature clustering strategy along with an adaptive pruning mechanism, enabling precise and effective watermark removal.

\item 
Extensive experiments across various 3DGS watermarking methods demonstrate that our approach consistently 
achieves superior watermark purification effectivess while preserving original scene fidelity.
\end{itemize}

%% file: sec/2_Relatedwork.tex
\section{Related Works} \label{sec:formatting}

\noindent \textbf{3D Gaussian Splatting.} The success of 3DGS lies in its discrete 3D Gaussian representation of scenes, which has led to a surge of research interest. Recent efforts have focused on enhancing rendering quality \cite{yu2024mip}, improving efficiency \cite{fan2024instantsplat,chen2024mvsplat}, and optimizing storage \cite{navaneet2024compgs}. Additionally, 3DGS has been extended to dynamic 3D scenes \cite{wu20234d,li2024st}, large-scale outdoor environments \cite{liu2024citygaussian}, and high-speed scenarios \cite{xiong2024event3dgs,yu2024evagaussians,guo2024spikegs}. Researchers have also relaxed its constraints on camera poses \cite{muller2022instant} and sharp images \cite{chen2024deblur}. Beyond reconstruction, 3DGS is widely applied in inverse rendering \cite{guo2024prtgs,liang2023gsir,gao2023relightable}, 3D generation \cite{yi2023gaussiandreamer}, and 3D editing \cite{chen2023gaussianeditor}. 


\noindent \textbf{Copyright Protection in 3D Gaussian Splatting.} Due to the high value of 3DGS assets, various copyright protection techniques have emerged to safeguard the ownership and authenticity. GS-Hider \cite{zhang2024gshider}, the first dedicated 3DGS steganography framework, replaces spherical harmonics coefficients with secure features and employs dual decoders to disentangle hidden messages from scene content. Splats in Splats \cite{guo2024splats} improves usability by embedding watermarks via importance-graded spherical harmonics encryption and opacity mapping, preserving vanilla 3DGS attributes. 
Finally, SecureGS \cite{zhang2025securegs} adopts a anchor-point design with hybrid Gaussian encryption, embedding hidden attributes into anchor features, achieving superior fidelity and speed.

\noindent \textbf{Watermark Purification.} Traditional 2D watermark removal methods target superficial perturbations (e.g., cropping, rotation, noise) or leverage neural networks for erasure  \cite{liu2021wdnet, cao2019generative}. However, the unique rendering pipeline of 3DGS, which optimizes millions of anisotropic Gaussians through differentiable splatting, embeds watermarks directly into the scene's geometric and photometric attributes. This fundamentally shifts the focus of watermark attacks: \textbf{instead of targeting the rendered image, adversaries must now operate directly on the 3DGS model itself.}
Our work pioneers this direction, proposing 3D-aware watermark purfication framework that bypass 2D constraints to challenge the robustness of 3DGS copyright protecting frameworks.

\begin{figure*}[t]
  \centering
  \includegraphics[width=0.98\linewidth]{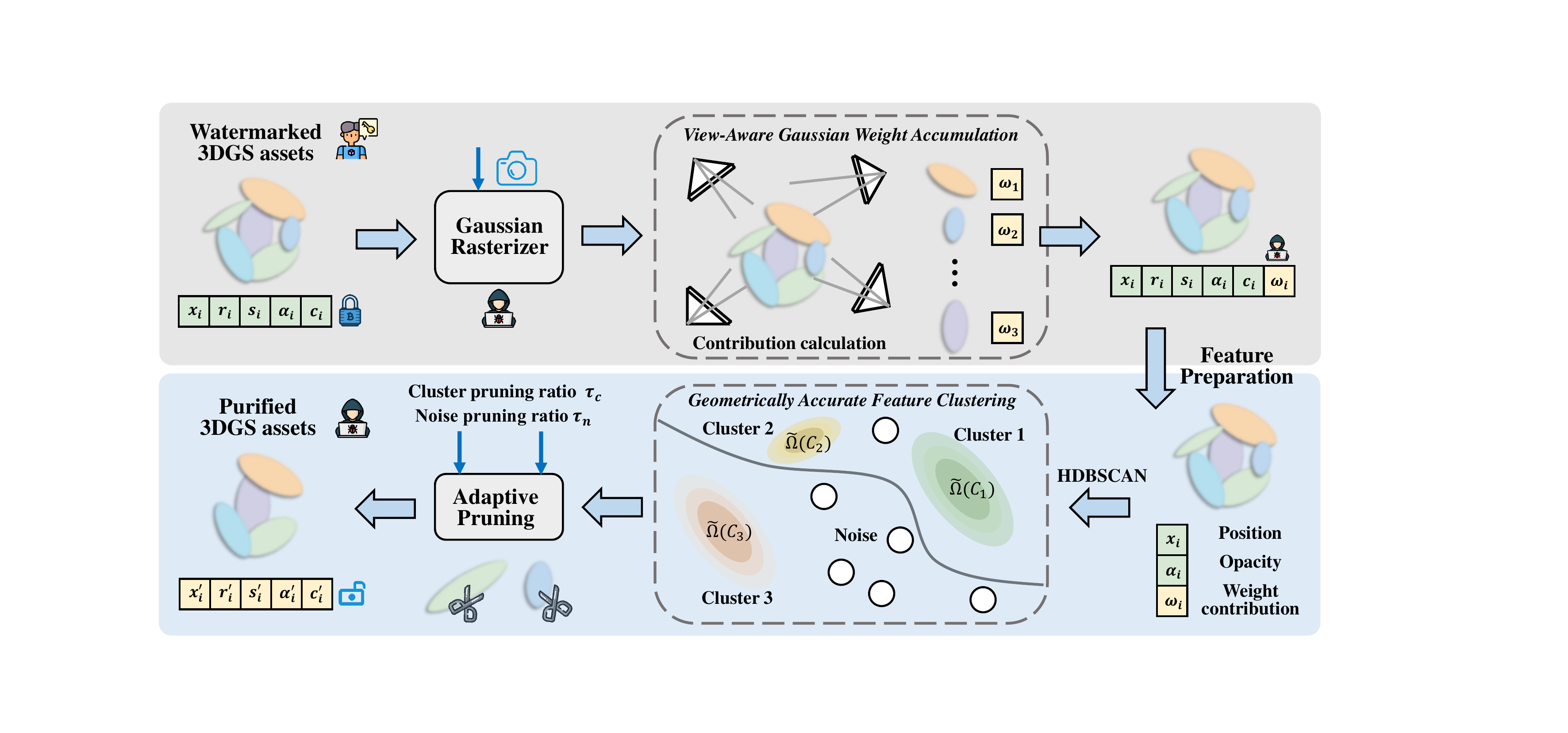}
  \caption{\textbf{Overview of our GSPure framework.} \textbf{First,} we compute the rendering contribution of each Gaussian primitive from multiple viewpoints.
  \textbf{Next,} we construct geometry-aware features and use HDBSCAN to identify watermark-correlated clusters. \textbf{Finally,} adaptive pruning based on cluster-level and noise-level thresholds ($\tau_c$ and $\tau_n$) effectively removes watermark-related Gaussians while preserving original scene fidelity.}
  \label{fig2}
\end{figure*}

%% file: sec/4_method.tex
\section{Methodology}
\subsection{Preliminaries}
3DGS provides an explicit and efficient representation of 3D scenes by modeling them as a collection of anisotropic Gaussian primitives. Each Gaussian primitive $\mathcal{G}$ is parameterized by its spatial position $x \in \mathbb{R}^3$ and covariance matrix $\Sigma \in \mathbb{R}^{3\times3}$, which defines its shape and orientation: 
\begin{align}
    \mathcal{G}(x) = \exp\left(-\frac{1}{2} (x - \mu)^T \Sigma^{-1} (x - \mu)\right).
\end{align}
Where $\mu$ represents the mean position, the covariance matrix $\Sigma$ is factorized into a rotation matrix $R$ and a diagonal scale matrix $S$, ensuring independent control over shape and orientation:
\begin{align}
    \Sigma = R S S^T R^T.
\end{align}
Rendering is achieved by projecting 3D Gaussians onto a 2D image plane. This transformation is performed by computing the projected covariance $\Sigma'$ in screen space:
\begin{align}
    \Sigma' = J W \Sigma W^T J^T,
\end{align}
where $J$ is the Jacobian of the projective transformation, and $W$ is the camera-view transformation matrix. After projection, the final image is obtained using alpha compositing, where Gaussians are blended in a front-to-back order:
\begin{align}
    C = \sum_{i} c_i \alpha_i \prod_{j=1}^{i-1} (1 - \alpha_j),
\end{align}
where $c_i$ represents the color of each Gaussian, and $\alpha_i$ is its opacity. Each Gaussian primitive is defined by a set of attributes:$\{x_i, r_i, s_i, \alpha_i, c_i\},$ where $x_i \in \mathbb{R}^3$ denotes position, $r_i \in \mathbb{R}^4$ encodes rotation as a quaternion, $s_i \in \mathbb{R}^3$ defines scale, $\alpha_i$ is a scalar opacity value, and $c_i \in \mathbb{R}^{n \times 3}$ represents the color.

\subsection{Problem Settings and Intuitive Methods} \label{3_2}
For \textit{\textbf{copyright owners}} of 3DGS assets, embedding watermarks in point cloud files allows for traceability when models are publicly shared. Unauthorized users who exploit these assets can be identified through watermark extraction. For the \textit{\textbf{copyright infringer}}, the goal is to directly purify watermarks from the 3DGS file while preserving its usability for downstream tasks such as rendering and editing. If successful, such an attack would fundamentally compromise existing 3DGS copyright protection frameworks, revealing critical vulnerabilities.

\textbf{Intuitive 3DGS watermark purification methods.} \textit{1) Random Gaussian Pruning:} Unlike image cropping, it removes randomly selected Gaussian primitives along with their attributes (e.g., opacity, color).  
\textit{2) Feature Scaling:} This approach scales the color and opacity attributes of Gaussian primitives, potentially weakening the visibility of watermark-related Gaussians.  
\textit{3) Attribute Noise Injection:} Adding noise to photometric features, such as spherical harmonics coefficients, aims to obfuscate watermark signals embedded in Gaussian parameters.  \textit{4) Surface-Aware Purification:} Extracting scene surfaces using Gaussian Opacity Fields (GOF) \cite{yu2024gaussian} attempts to isolate core geometry from watermark-related artifacts.  While these methods introduce perturbations to Gaussian parameters, the inherent coupling and redundancy of 3DGS watermarking significantly limit their effectiveness. These approaches either fail to remove the watermark completely or simultaneously degrade the fidelity of the original scene, making them unsuitable for precise watermark purification.

\subsection{The Proposed Purification Framework GSPure}
\textbf{Motivation and overview.} As demonstrated above, existing 3DGS watermarking frameworks either exhibit tight coupling between watermark information and the original scene \cite{zhang2024gshider,zhang2025securegs}, or rely on hidden watermark recovery viewpoints and attributes unknown to the attacker \cite{guo2024splats}. This makes traditional watermark attacks based on simple cropping, noise injection, or filtering ineffective. However, we observe that watermarks embedded through parametric optimization exhibit strong correlations with highly related Gaussian primitives in terms of position, opacity, and their role within the rendering pipeline. Specifically, Gaussian primitives that play critical roles in rendering the original scene often remain inactive during watermark recovery. This indicates that analyzing Gaussian contributions from different viewpoints in the original, user-visible scene presents an opportunity to isolate Gaussian primitives highly correlated with the watermark. Inspired by this insight, we propose a \textbf{\textit{view-aware gaussian weight accumulation}} method to quantify the contribution weight of each Gaussian under various viewpoints. Furthermore, due to viewpoint discontinuities and the strong coupling between the original and watermark scenes, relying solely on weighted contributions could partially damage the original scene or leave some watermark information intact. Thus, we further propose a \textbf{\textit{geometrically accurate feature clustering}} strategy, combining Gaussian positions, opacity, and weighted contributions as joint features. By applying pruning at the cluster level, our method achieves a more effective separation between watermarks and original scenes.

\textbf{View-Aware Gaussian Weight Accumulation.} We represent the publicly available point cloud as a set of Gaussian primitives $\{\mathcal{G}_k\ |\ k = 1, \dots, K\}$ in $\mathbb{R}^3$. Each Gaussian $\mathcal{G}_k$ is parameterized by its center position $\mathbf{p}_k \in \mathbb{R}^3$, opacity $\alpha_k$, rotation matrix $\mathbf{R}_k \in \mathbb{R}^{3\times3}$, and scale matrix $\mathbf{S}_k \in \mathbb{R}^{3\times3}$, defined explicitly as:
\begin{equation}
    \mathcal{G}_k(\mathbf{x}) = \exp\left(-\frac{1}{2}(\mathbf{x}-\mathbf{p}_k)^T \Sigma_k^{-1}(\mathbf{x}-\mathbf{p}_k)\right),
\end{equation}
where $\Sigma_k = \mathbf{R}_k\mathbf{S}_k\mathbf{S}_k^\top\mathbf{R}_k^\top$ is the covariance matrix of Gaussian $\mathcal{G}_k$. For a specific rendering viewpoint $v$, given a camera center $\mathbf{o}_v$ and a ray direction $\mathbf{r}_v$, let the ray intersect with the Gaussian $\mathcal{G}_k$ at a distance $t_v$ along the ray, such that the intersection point is $\mathbf{x}_v = \mathbf{o}_v + t_v\mathbf{r}_v$. We then define the ray-Gaussian intersection energy $\mathcal{E}(\mathcal{G}_k, \mathbf{o}_v, \mathbf{r}_v)$ as follows:
\begin{equation}
\begin{split}
\mathcal{E}(\mathcal{G}_k, \mathbf{o}_v, \mathbf{r}_v) 
= \exp\Big( -\frac{1}{2} \big(
    \mathbf{r}_v^T \Sigma_k^{-1} \mathbf{r}_v t_v^2 
    + \\ 2\mathbf{o}_v^T \Sigma_k^{-1} \mathbf{r}_v t_v 
    + \mathbf{o}_v^T \Sigma_k^{-1} \mathbf{o}_v
\big) \Big),
\end{split}
\end{equation}
where $t_v$ is the intersection distance from the center $\mathbf{o}_v$ along the direction $\mathbf{r}_v$ to the Gaussian $\mathcal{G}_k$.

Considering occlusion effects, the contribution of Gaussian $\mathcal{G}_k$ at viewpoint $v$ is modulated by the opacity and intersection energy of preceding Gaussians following an alpha-blending mechanism. We denote this viewpoint-dependent contribution as $\omega_{k,v}$:
\begin{equation}
\omega_{k,v} = \alpha_k \cdot \mathcal{E}(\mathcal{G}_k, \mathbf{o}_v, \mathbf{r}_v)\prod_{j=1}^{k-1}(1 - \alpha_j\mathcal{E}(\mathcal{G}_j, \mathbf{o}_v, \mathbf{r}_v)).
\end{equation}
Where $\alpha_j$ denotes the opacity, indicating the rendering order, and the product term represents cumulative transmittance due to occlusion by previous Gaussians. To capture the overall importance of each Gaussian across multiple views, we define a \textit{view-accumulated weight} $\omega_k$ by averaging these rendering contributions across $N$ rendered viewpoints: $\omega_k = \frac{1}{N}\sum_{v=1}^{N}\omega_{k,v}.$ Intuitively, the weight $\omega_k$ quantifies the \textbf{average rendering contribution} of Gaussian $\mathcal{G}_k$ across different viewpoints. Gaussians highly correlated with watermark information typically exhibit lower $\omega_k$ due to their inconsistent or viewpoint-specific rendering behaviors. Conversely, Gaussians strongly associated with the original scene demonstrate higher and more consistent contribution values across viewpoints.

\begin{figure*}[t]
  \centering
  \includegraphics[width=0.9\linewidth]{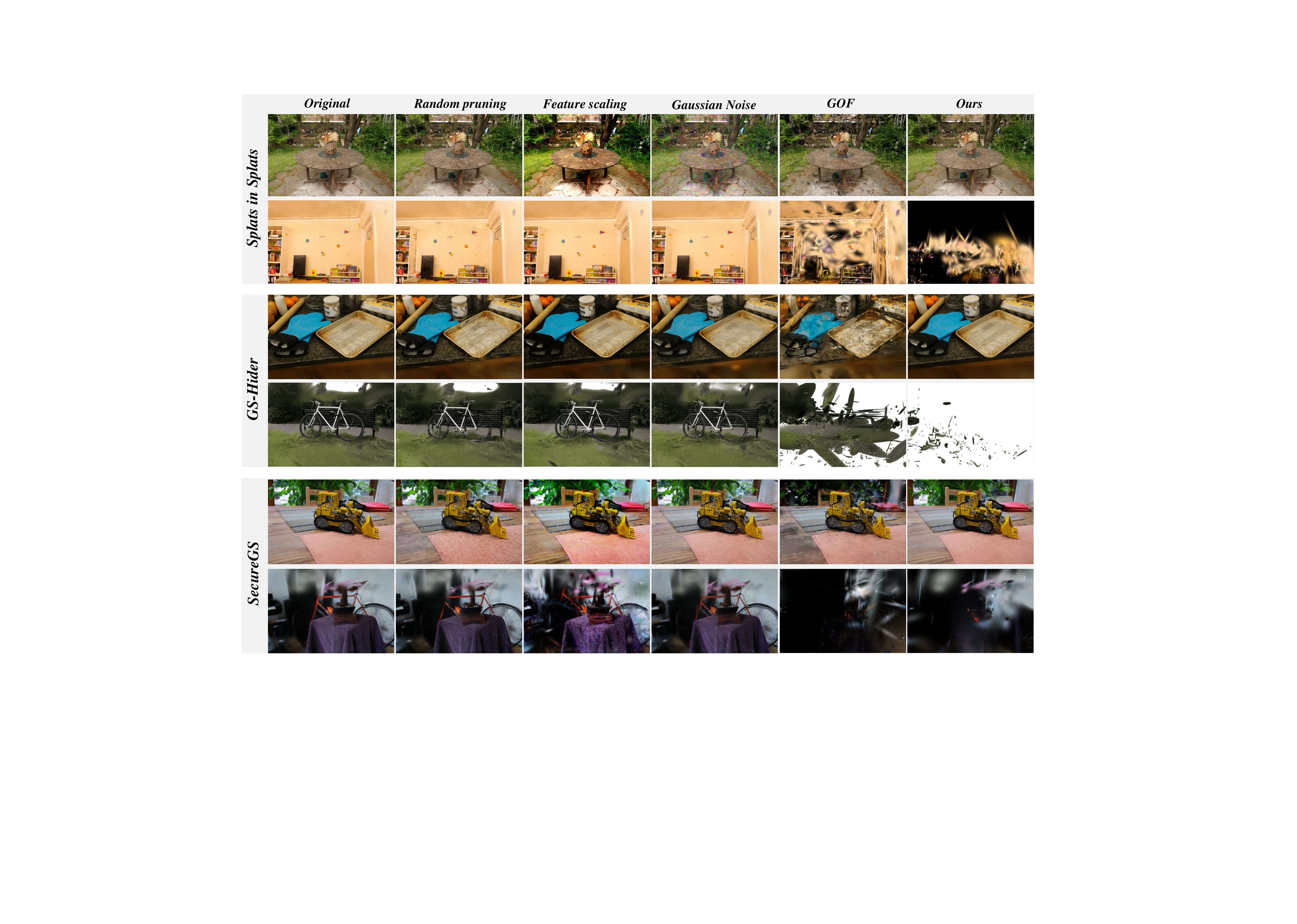}
  \caption{Qualitative comparisons on Mipnerf360 datasets. The first row of each group represents the \textbf{original scene} while the second represents the \textbf{hidden watermark}. Our GSPure effectively removes watermarks across all 3DGS watermarking methods while preserving the integrity of the original scene with \textbf{minimal impact}.
}
  \label{fig_compare}
\end{figure*}

\textbf{Geometrically Accurate Feature Clustering.}
Due to the training manner of scene hiding and the density mechanism of Gaussian primitives, watermark-related Gaussian primitives typically exhibit strong geometric and opacity correlations \cite{guo2024splats}. Inspired by this observation, we combine the computed view-accumulated weights with Gaussian positional and opacity information to construct a unified \textit{geometrically coherent feature}, which effectively mitigates discontinuities across viewpoints and breaks the tight coupling between watermark information and the original scene. Specifically, we first standardize Gaussian positions, opacities, and view-accumulated weights individually, and then concatenate these standardized values into the following high-dimensional feature vector $\mathbf{F}_k$:
\begin{equation}
    \mathbf{F}_k = \text{Concat}\left(\frac{\mathbf{p}_k - \mu_p}{\sigma_p}, \frac{\alpha_k - \mu_\alpha}{\sigma_\alpha}, \frac{\omega_k - \mu_\omega}{\sigma_\omega}\right),
\end{equation}
where $\mu_p/\sigma_p$, $\mu_\alpha/\sigma_\alpha$, and $\mu_\omega/\sigma_\omega$ denote the mean and standard deviation for positions, opacities, and weights computed across all Gaussian primitives, respectively. 

Considering the significant variability of Gaussian primitives across different scenes, we employ the adaptive density-based hierarchical clustering method HDBSCAN~\cite{mcinnes2017hdbscan} to effectively capture diverse distributions. This yields a collection of coherent clusters along with noise points, defined as: $  \mathcal{C} = \{C_1, ..., C_M\} \cup \mathcal{N},$ where $C_i$ denotes a cluster and $\mathcal{N}$ denotes the noise points.

To quantify the importance of each cluster, we compute the average weight contribution $\widetilde{\Omega}(C_i)$ of each cluster $C_i$ by averaging the view-accumulated weights:
\begin{equation}
    \widetilde{\Omega}(C_i) = \frac{1}{|C_i|}\sum_{k \in C_i} \omega_k.
\end{equation}

\textbf{Adaptive Pruning.} For pruning, we consider that Gaussian primitives within the same cluster contribute consistently. Consequently, we prune at the cluster level: if the average contribution of a cluster falls below a dynamically determined threshold, all Gaussians within the cluster are removed. Conversely, noise points are evaluated individually. Specifically, watermark-related Gaussian clusters typically exhibit a significant gap in average contribution compared to the global mean, prompting us to introduce two adaptive pruning thresholds: a cluster pruning factor $\tau_c$ and a noise pruning factor $\tau_n$. These factors dynamically adjust pruning thresholds based on the global average Gaussian contribution $\bar{\omega}= \frac{1}{K}\sum_{i=1}^{K}\omega_{i}$, ensuring selective removal while preserving essential scene structures. Our pruning mechanism is thus expressed as follows:
\begin{equation}
    Prune(\mathcal{G}_k) =
    \begin{cases}
        \mathbb{I}\left(\widetilde{\Omega}(C_i) < \frac{\bar{\omega}}{\tau_c}\right), & \text{if } \mathcal{G}_k \in C_i,\\[6pt]
        \mathbb{I}\left(w_k < \frac{\bar{\omega}}{\tau_n}\right), & \text{if } \mathcal{G}_k \in \mathcal{N},
    \end{cases}
 \label{prune}
\end{equation}
where $\mathbb{I}(\cdot)$ is the indicator function, and thresholds $\tau_c$ and $\tau_n$ dynamically adjust the pruning criteria based on the global average gaussian weight $\bar{\omega}$. This ensures the flexibility of our framework, achieving effective watermark purification while preserving original scene integrity.

%% file: sec/5_Experiment.tex
\begin{table*}[t]
\setlength{\tabcolsep}{1.0pt} 
\renewcommand{\arraystretch}{1.25}
\centering
\resizebox{\textwidth}{!}{
\begin{tabular}{c||cccccccccccccccccc|c}
\hline\thickhline
\rowcolor{gray!10}
& \multicolumn{2}{c}{Bicycle} & \multicolumn{2}{c}{Bonsai} & \multicolumn{2}{c}{Room} & \multicolumn{2}{c}{Flowers} & \multicolumn{2}{c}{Treehill} & \multicolumn{2}{c}{Garden} & \multicolumn{2}{c}{Stump} & \multicolumn{2}{c}{Counter} & \multicolumn{2}{c|}{Kitchen} & \\
 
\rowcolor{gray!10}
\multirow{-2}{*}{Method} & Scene & Message & Scene & Message & Scene & Message & Scene & Message & Scene & Message & Scene & Message & Scene & Message & Scene & Message & Scene & Message & \multirow{-2}{*}{Score} \\
\hline\hline
\textbf{\textit{Splats in Splats }}
& 24.43 & 28.99 & 30.95 & 25.83 & 30.93 & 23.36 & 20.69 & 29.01 & 22.16 & 22.79 & 26.83 & 29.01 & 25.41 & 28.81 & 28.64 & 22.52 & 30.70 & 28.35 & 0.00\\

Random Pruning
& 22.35 & 26.96 & 26.31 & 23.68 & 27.31 & 20.78 & 18.85 & 26.99 & 20.72 & 20.73 & 23.75 & 26.72 & 23.24 & 26.89 & 25.60 & 20.71 & 24.97 & 25.51 & -0.88\\

Feature scaling
& 18.03 & 29.33 & 18.83 & 26.36 & 18.43 & 24.14 & 16.47 & 29.42 & 17.21 & 23.24 & 17.72 & 29.46 & 17.90 & 29.29 & 18.01 & 22.93 & 18.24 & 29.10 & -9.38\\

Gaussian Noise
& 19.90 & 29.33 & 21.08 & 26.36 & 21.01 & 24.14 & 17.93 & 29.42 & 18.64 & 23.24 & 20.55 & 29.46 & 20.23 & 29.29 & 20.75 & 22.93 & 20.81 & 29.10 & -7.15\\

GOF 
& 18.56 & 16.66 & 18.43 & 13.55 & 19.47 & 12.60 & 14.58 & 18.34 & 17.09 & 17.87 & 19.34 & 12.89 & 19.70 & 20.44 & 16.58 & 11.79 & 19.05 & 16.50 & 2.24\\

\rowcolor{black!15} Ours
& \textbf{23.20} & \textbf{8.27} & \textbf{26.48} & \textbf{9.33} & \textbf{30.24} & \textbf{8.19} & \textbf{19.54} & \textbf{11.13} & \textbf{21.40} & \textbf{13.84} & \textbf{25.10} & \textbf{9.42} & \textbf{23.88} & \textbf{9.30} & \textbf{28.60} & \textbf{10.14} & \textbf{27.59} & \textbf{10.49} & \textbf{15.21}\\

\hline
\textbf{\textit{GS-Hider }}
& 23.20 & 19.43 & 31.06 & 25.28 & 30.94 & 23.00 & 20.36 & 23.08 & 19.59 & 20.51 & 25.11 & 24.68 & 24.84 & 25.00 & 26.97 & 20.48 & 29.80 & 24.35 & 0.00\\

Random Pruning
& 21.25 & 15.40 & 27.19 & 21.31 & 28.39 & 20.17 & 18.37 & 19.30 & 18.51 & 18.98 & 22.00 & 21.29 & 22.99 & 21.73 & 24.79 & 18.93 & 25.22 & 19.51 & 0.68\\

Feature scaling
& 20.64 & 17.11 & 26.04 & 23.48 & 24.36 & 21.46 & 19.26 & 22.44 & 17.84 & 19.32 & 24.08 & 23.45 & 23.77 & 23.89 & 22.68 & 19.84 & 27.49 & 21.99 & -1.42\\

Gaussian Noise
& 22.68 & 17.10 & 25.90 & 23.32 & 27.56 & 20.43 & 19.36 & 22.05 & 19.20 & 20.51 & 23.68 & 23.15 & 22.07 & 21.60 & 25.60 & 19.92 & 27.78 & 18.78 & 0.11\\

GOF 
& 11.91 & 11.42 & 17.61 & 10.96 & 18.96 & 14.71 & 11.56 & 14.15 & 14.39 & 16.25 & 14.05 & 13.71 & 14.97 & 14.22 & 17.60 & 10.42 & 15.30 & 12.83 & -0.93 \\

\rowcolor{black!15} Ours
& \textbf{23.11} & 11.60 & \textbf{29.55} & \textbf{9.76} & \textbf{30.06} & \textbf{13.95} & \textbf{20.25} & 14.18 & \textbf{19.48} & \textbf{14.56} & \textbf{24.91} & 13.88 & \textbf{24.69} & 14.26 & \textbf{26.17} & \textbf{3.29} & 27.49 & 12.83 & \textbf{10.16}\\

\hline
\textbf{\textit{SecureGS }}
& 24.24 & 21.80 & 30.50 & 21.37 & 31.22 & 21.58 & 20.57 & 23.84 & 22.22 & 23.78 & 26.62 & 26.88 & 25.52 & 24.62 & 28.50 & 20.85 & 29.78 & 23.24 & 0.00\\

Random Pruning
& 21.60 & 20.27 & 24.89 & 19.45 & 26.78 & 19.61 & 18.60 & 21.54 & 20.51 & 21.55 & 22.54 & 24.40 & 22.85 & 22.52 & 24.99 & 19.25 & 23.30 & 20.71 & -1.59 \\

Feature scaling
& 19.30 & 16.71 & 21.96 & 13.09 & 22.60 & 14.19 & 17.00 & 16.79 & 17.17 & 17.37 & 20.31 & 18.80 & 20.24 & 17.45 & 20.70 & 16.39 & 21.73 & 15.60 & 0.39\\

Gaussian Noise
& 23.62 & 21.63 & 28.52 & 20.51 & 30.24 & 20.94 & 20.08 & 23.37 & 21.82 & 23.27 & 25.64 & 26.55 & 24.89 & 24.18 & 26.92 & 20.55 & 27.99 & 22.57 & -0.55\\

GOF
& 17.57 & 11.49 & 14.21 & 11.67 & 14.47 & 9.83 & 12.13 & 11.51 & 12.69 & 13.29 & 17.16 & 10.80 & 17.87 & 10.33 & 15.34 & 11.37 & 17.08 & 11.28 & 0.64 \\

\rowcolor{black!15} Ours
& 23.44 & 15.30 & 26.81 & 14.81 & \textbf{30.35} & 20.25 & 19.34 & 16.10 & 21.38 & 15.79 & 26.29 & 20.93 & 24.52 & 17.35 & \textbf{27.55} & 15.13 & 27.73 & 14.77 & \textbf{5.03}\\

\hline\thickhline
\end{tabular}
    }
\caption{
Quantitative PSNR comparison on Mip-NeRF 360 dataset. Bold values indicate the best performance. The ``Score" metric summarizes overall effectiveness, balancing watermark removal performance with minimal scene disturbance.
}
\label{tab_detailed_performance}
\end{table*}
\section{Experiment}
\subsection{Experimental Settings}

\textbf{Implementation Details.} 
We adopt the training parameters provided in Splats in Splats, GS-Hider, and SecureGS, respectively. All experiments are conducted on an NVIDIA A800 GPU. Unless otherwise stated, the thresholds $\tau_n$ and $\tau_c$ in Eq.~\ref{prune} are set to $(\tau_n, \tau_c) = (4,4)$ for Splats in Splats, $(4,4)$ for GS-Hider, and $(2,3)$ for SecureGS.  For SecureGS, all weight analyses are performed at the anchor level. Additionally, we modify the CUDA implementation of 3DGS to support the computation of rendering contribution weights.

\subsection{Experimental Settings}

\textbf{Datasets.} 
Following existing 3DGS watermarking methods~\cite{guo2024splats,zhang2024gshider,zhang2025securegs}, we conduct main experiments on Mip-NeRF360 dataset~\cite{barron2022mipnerf360}. We maintain consistent scene-watermark associations used by GS-Hider~\cite{zhang2024gshider}. 

\textbf{Baselines.} 
We adapt traditional 2D watermark purification techniques for comparative evaluation. We select four baseline methods: (1) \textit{Random Gaussian Pruning}, where a certain ratio of Gaussian primitives are randomly removed; (2) \textit{Feature Scaling}, which amplifies Gaussian color features (or anchor features in SecureGS); (3) \textit{Gaussian Noise Injection}, where Gaussian noise perturbations are introduced to attributes of Gaussian primitives; and (4) \textit{Gaussian Opacity Fields (GOF)}~\cite{yu2024gaussian}, which explicitly extracts the surface of the original scene. 

\textbf{Metrics.}
We evaluate the fidelity of Novel View Synthesis of original scenes using Peak Signal-to-Noise Ratio (PSNR) and Structural Similarity Index Measure (SSIM). Additionally, we introduce a \textit{Score} metric to comprehensively assess watermark removal effectiveness, defined as:
\begin{equation}
    Score = \Delta PSNR_{message} - \Delta PSNR_{scene}.
\end{equation}
A higher \textit{Score} indicates more effective watermark removal with better preservation of the original scene fidelity.

\begin{table*}[h]
  \centering
  \begin{minipage}{0.48\linewidth}
    \centering
    \resizebox{\linewidth}{!}{
    \begin{tabular}{ccc||cccc}
     \hline\thickhline
     \rowcolor{gray!10}  &  &   &  \multicolumn{2}{c}{GS-Hider} & \multicolumn{2}{c}{SecureGS } \\
     \rowcolor{gray!10} \multirow{-2}{*}{Opacity}& \multirow{-2}{*}{Weight} &  \multirow{-2}{*}{Clustering}  & PSNR$\downarrow$ & SSIM$\downarrow$ & PSNR$\downarrow$ & SSIM$\downarrow$ \\
     \hline\hline
     \Checkmark & \XSolidBrush & \XSolidBrush & 14.67 & 0.55 & 21.04 & 0.69 \\
     \XSolidBrush & \Checkmark & \XSolidBrush  & 14.20 & 0.53 & 19.41 & 0.64 \\
     \Checkmark & \Checkmark & \XSolidBrush & 12.29 & 0.49 & 20.27 & 0.67 \\
     \Checkmark & \XSolidBrush & \Checkmark & 12.81 & 0.54 & 17.61 & 0.61 \\
     \XSolidBrush & \Checkmark & \Checkmark  & 12.82 & 0.54 & 16.88 & 0.59 \\
   \rowcolor{black!15} \Checkmark & \Checkmark & \Checkmark & \textbf{12.03} & \textbf{0.49} & \textbf{16.71} & \textbf{0.56} \\
     \hline\thickhline
    \end{tabular}
    }
    \caption{Quantitative comparison of watermark residuals under GS-Hider and SecureGS.}
    \label{table:configurations1}
  \end{minipage}
  \hfill
  \begin{minipage}{0.48\linewidth}
    \centering
    \resizebox{\linewidth}{!}{
    \begin{tabular}{ccc||cccc}
     \hline\thickhline
     \rowcolor{gray!10}  &  &  & \multicolumn{2}{c}{Scene} &  \multicolumn{2}{c}{Message}  \\
     \rowcolor{gray!10} \multirow{-2}{*}{Opacity} & \multirow{-2}{*}{Weight} &  \multirow{-2}{*}{Clustering} & PSNR$\uparrow$ & SSIM$\uparrow$ & PSNR$\downarrow$ & SSIM$\downarrow$ \\
     \hline\hline
     \Checkmark & \XSolidBrush & \XSolidBrush & 26.25 & 0.77 & 21.25 & 0.70 \\
     \XSolidBrush & \Checkmark & \XSolidBrush & 24.33 & 0.73 & 9.60 & 0.27 \\
     \Checkmark & \Checkmark & \XSolidBrush & 24.26 & 0.71 & 9.52 & 0.26 \\
     \Checkmark & \XSolidBrush & \Checkmark & 22.47 & 0.65 & \textbf{9.06} & \textbf{0.21} \\
     \XSolidBrush & \Checkmark & \Checkmark & 25.00 & 0.76 & 15.76 & 0.52 \\
    \rowcolor{black!15} \Checkmark & \Checkmark & \Checkmark & \textbf{25.62} & \textbf{0.77} & 10.19 & 0.28 \\
     \hline\thickhline
    \end{tabular}
    }
    \caption{Quantitative comparisons of scene fidelity and watermark residual under Splats in Splats.}
    \label{table:configurations2}
  \end{minipage}
\end{table*}

\subsection{Main Results}
We evaluate the effectiveness of GSPure against four baselines across three 3DGS watermarking frameworks (Splats in Splats~\cite{guo2024splats}, GS-Hider~\cite{zhang2024gshider}, and SecureGS~\cite{zhang2025securegs}). Comprehensive results summarized in Fig.~\ref{fig_compare} and Tab.~\ref{tab_detailed_performance} demonstrate that GSPure consistently achieves superior watermark purification while retaining the fidelity of original scenes across multiple metrics, validating its efficacy and transferability.

\textbf{Original Scene Fidelity.} 
Compared to baselines, GSPure effectively preserves the fidelity of the original scene while removing watermarks. As shown in Fig.~\ref{fig_compare}, the rendered results of GSPure after pruning are nearly indistinguishable from the original scene, demonstrating the robustness of our weight computation and clustering strategy in decoupling watermark-related Gaussians from those contributing to the original scene. For instance, under \textit{Splats in Splats}, GOF, Feature Scaling, and Gaussian Noise significantly degrade both visual and geometric fidelity. Similar degradation is observed across the other two watermarking methods. The quantitative results in Tab.~\ref{tab_detailed_performance} further verify this observation. In the \textit{GS-Hider} framework, our pruning approach reduces the PSNR of the original scene by only \textbf{0.68} on average, substantially outperforming GOF and Gaussian Noise, which cause PSNR drops of \textbf{9.68} and \textbf{2.21}, respectively. 

Notably, under \textit{SecureGS}, GSPure does not achieve the highest scene fidelity, yielding a PSNR of $25.27$, slightly lower than Gaussian Noise ($25.53$). This is due to SecureGS’s reliance on the Scaffold-GS anchor-based design, which inherently provides robustness against Gaussian noise perturbations. However, this also results in a major drawback: Gaussian Noise fails to effectively remove watermarks, demonstrating nearly no impact. Overall, GSPure exhibits superior performance in preserving scene fidelity.

\textbf{Watermark Purification Effectiveness.} GSPure significantly outperforms all baselines in watermark purification. As illustrated in Fig.~\ref{fig_compare}, the extracted watermark after applying GSPure contains minimal residual information from the original watermark. under both \textit{GS-Hider} and \textit{Splats in Splats}, nearly no watermark traces remain, whereas methods like Random Pruning struggle to effectively purify watermarks. Notably, GOF \cite{yu2024gaussian} achieves relatively strong watermark removal, even surpassing GSPure in \textit{SecureGS}. However, this comes at a severe cost: GOF causes irreversible and unacceptable damage to the original scene. 

The quantitative results in Tab.~\ref{tab_detailed_performance} reinforce these findings. Under \textit{Splats in Splats}, GSPure achieves the most significant improvement, surpassing the second-best method by $5.44$ PSNR in watermark removal. Additionally, GSPure achieves the highest \textit{Score} across all three watermarking methods, consistently outperforming the second-best approach by margins of $12.97$, $9.48$, and $4.39$. This consistently strong performance underscores the effectiveness of our weight-based adaptive clustering mechanism, which selectively prunes watermark-related Gaussian primitives while preserving the integrity of the original visual content.

\subsection{Ablation Study}
\noindent \textbf{Necessity of Weight Accumulation and Feature Clustering.} 
To thoroughly evaluate the effectiveness of two core components, we conduct an ablation study with different configurations involving opacity filtering (Opacity), weight contribution analysis (Weight), and clustering (Cluster). Each configuration represents a variant of GSPure. The results in Tab.~\ref{table:configurations1} and Tab.~\ref{table:configurations2} indicate that the fully configured GSPure achieves the best overall performance in most cases. For instance, on GS-Hider, it reduces the watermark PSNR to \textbf{12.03} and SSIM to \textbf{0.49}, demonstrating its effectiveness in watermark removal. In contrast, on Splats in Splats, using only Weight or combining Opacity+Cluster achieves competitive watermark purification, even surpassing GSPure. However, these methods exhibit poor scene fidelity preservation, resulting in a lower overall score. Even though using Opacity+Weight or Weight+Cluster independently still delivers strong performance, we recommend the full GSPure configuration for the best generalization and robustness across diverse 3DGS watermarking methods.

\begin{figure}[h]
  \centering
  \includegraphics[width=\linewidth]{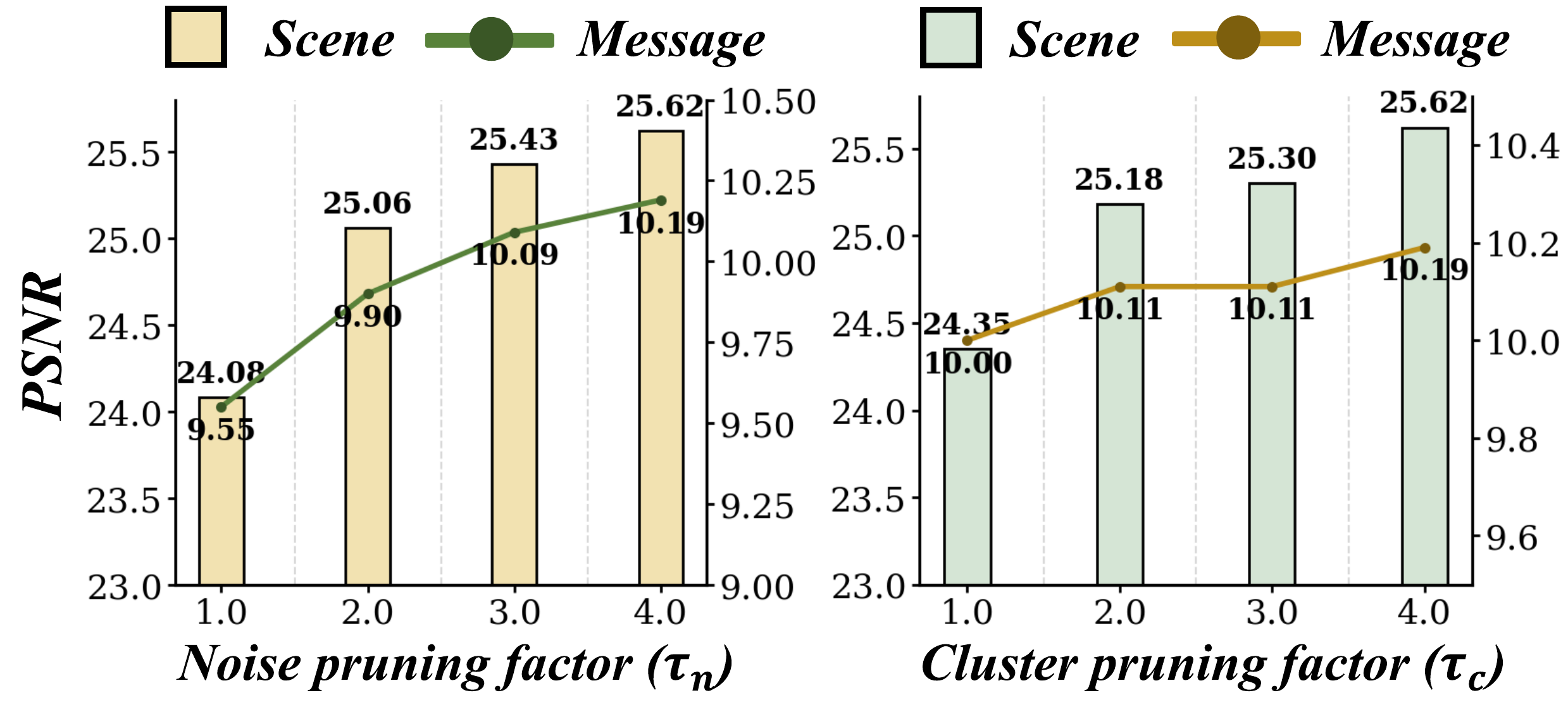}
  \caption{Impact of noise and cluster pruning factors on scene quality and watermark removal.}
  \label{fig_ab}
\end{figure}


\noindent\textbf{Pruning Thresholds.}  
Fig.~\ref{fig_ab} evaluates the impact of varying $\tau_n$ and $\tau_c$ from 1.0 to 4.0 on Splats in Splats. Overall, bigger $\tau_n$ will increase PSNR for both the scene and the watermark, aligning with our intuition that watermark-related Gaussian primitives in noise are highly entangled with normal scene primitives. Regarding $\tau_c$, we find that increasing its value has minimal impact on watermark removal but significantly improves scene quality. This is attributed to the effectiveness of feature clustering, ensuring that watermark-correlated Gaussian clusters exhibit a substantial gap from the pruning threshold. Based on these findings, we recommend using a relatively high $\tau_c$ and adjusting $\tau_n$ flexibly to achieve optimal purification performance.




\subsection{Visualization of Point Cloud}
\begin{figure}[h]
  \centering
    \includegraphics[width=\linewidth]{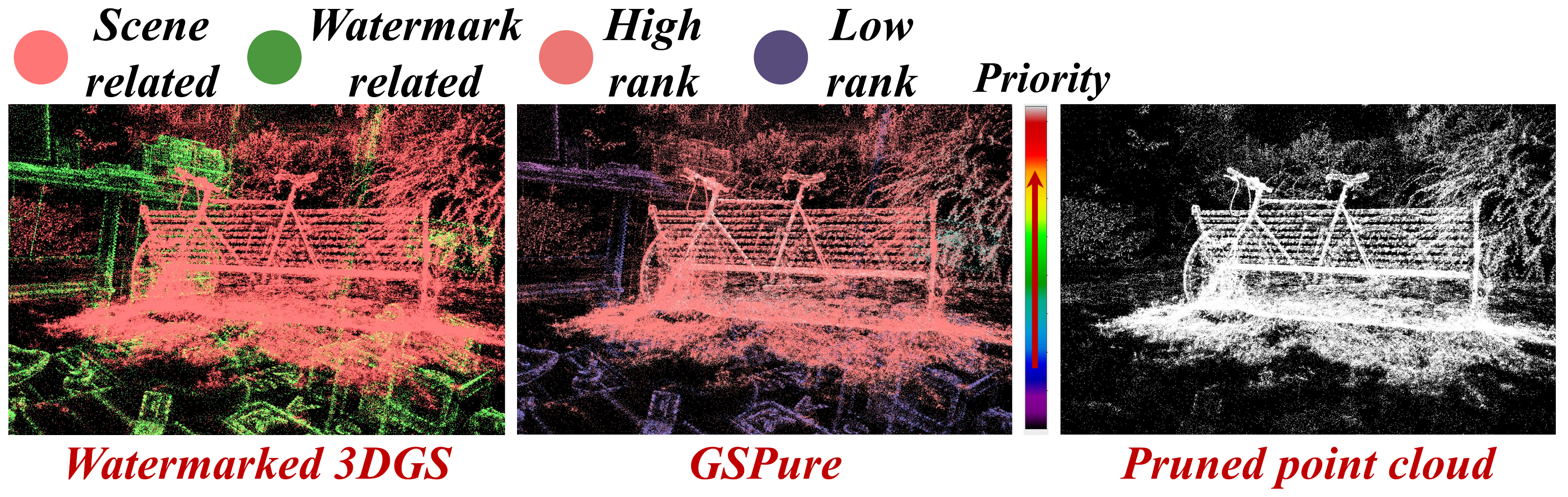}
\caption{Visualization of Point Cloud. The left column represents the ``Ground Truth''. The middle column shows the clustering results of our GSPure. The right column presents the pruned point cloud after GSPure.}
    \label{fig_ab_vis}
\end{figure}

From the perspective of copyright owners, we extract watermark-correlated and scene-correlated Gaussian primitives, as illustrated on the left side of Fig.~\ref{fig_ab_vis}. We then compute the Gaussian contribution weights and visualize the clustering results of GSPure in the middle column of Fig.~\ref{fig_ab_vis}. Surprisingly, GSPure nearly perfectly disentangles watermark-related Gaussian from those associated with the original scene, further validating the effectiveness of GSPure. Additionally, we observe that watermark-related Gaussian primitives tend to be clustered together, significantly increasing their likelihood of being pruned. The right column  presents the point cloud after pruning, demonstrating GSPure’s ability to maintain the fidelity of the original scene while achieving highly effective watermark removal.

\section{Conclusion}
In this work, we conduct the first systematic exploration of watermark purification for 3DGS watermarking and propose \textit{GSPure}, a novel framework specifically designed for effective watermark removal in 3DGS models. Unlike existing approaches that suffer from tight entanglement between watermark signals and original scene attributes, GSPure leverages a view-aware weight accumulation method alongside geometrically accurate feature clustering to precisely isolate and eliminate watermark-correlated Gaussian primitives. Our results demonstrate that current 3DGS watermarking methods do not provide absolute protection, as watermarks can be effectively removed while preserving the integrity of the original scene. We advocate for the development of more robust 3D Gaussian watermarking techniques and, in parallel, the advancement of principled and adaptive watermark removal methods. A deeper understanding of 3DGS watermark vulnerabilities and defenses is essential to fostering future research in copyright protection and adversarial robustness for 3D scene representations.